\documentclass[10pt,a4paper]{article}
\usepackage{hyperref}
\usepackage{lrec}

\usepackage{times}
\usepackage{latexsym}
\usepackage{booktabs}
\usepackage{CJKutf8}
\usepackage[whole,autotilde,substmingoth]{bxcjkjatype}
\usepackage{multirow}
\usepackage{array}
\usepackage[pdftex]{graphicx}

\usepackage{multibib}
\newcites{languageresource}{Language Resources}
\usepackage{tabularx}
\usepackage{soul}
\usepackage{epstopdf}
\usepackage{xstring}

\title{Construction of a Japanese Word Similarity Dataset}

\name{Yuya Sakaizawa\thanks{Currently at JustSystems Corporation.} and Mamoru Komachi}

\address{Tokyo Metropolitan University\\
	    6-6 Asahigaoka\\
	    Hino city, Tokyo 191-0065, Japan\\
	    {\tt \{sakaizawa-yuya@ed.,komachi@\}tmu.ac.jp}}

\abstract{
 An evaluation of distributed word representation is generally conducted using a word similarity task and/or a word analogy task.
 There are many datasets readily available for these tasks in English.
 However, evaluating distributed representation in languages that do not have such resources (e.g., Japanese) is difficult.
 Therefore, as a first step toward evaluating distributed representations in Japanese, we constructed a Japanese word similarity dataset.
 To the best of our knowledge, our dataset is the first resource that can be used to evaluate distributed representations in Japanese.
 Moreover, our dataset contains various parts of speech and includes rare
 words in addition to common words. \\ \newline 
}

\begin{document}

\maketitleabstract

\begin{figure*}[htbp]
	\begin{center}
		\newcolumntype{Y}{>{\centering\arraybackslash}p{22mm}} 
			\begin{tabular}{|Y|Y|Y|Y|Y|Y|Y|Y|} \hline
			\multirow{2}{15mm}{Sentence} & \multicolumn{5}{c|}{I don't think it is likely to not include these people, or [exclude]} \\ 
				 & \multicolumn{5}{c|}{\begin{CJK}{UTF8}{min}まさかこういった方々を対象としない、[排除する]わけではないと思いますが\end{CJK}} \\  \hline 
			\multirow{2}{15mm}{Paraphrase} & ignore & ostracize & avoid & exclude & remove  \\
				 & \begin{CJK}{UTF8}{min}無視する\end{CJK} & \begin{CJK}{UTF8}{min}排斥する\end{CJK} & \begin{CJK}{UTF8}{min}敬遠する\end{CJK} & \begin{CJK}{UTF8}{min}排除する\end{CJK} & \begin{CJK}{UTF8}{min}除外する\end{CJK}  \\ \hline
			 \end{tabular}
		 \caption{An example of the dataset from a previous study
         \citelanguageresource{kodaira}.}
		\label{tab:example}
	\end{center}
\end{figure*}

\section{Introduction}

Traditionally, a word is represented as a sparse vector indicating the word itself (one-hot vector) or the context of the word (distributional vector).
However, both the one-hot notation and  distributional notation suffer from data sparseness since dimensions of the word vector do not interact with each other.
Distributed word representation addresses the data sparseness problem by constructing a dense vector of a fixed length, wherein contexts are shared (or distributed) across dimensions.
Distributed word representation is known to improve the performance of many NLP applications such as machine translation \cite{chen-guo:2015:ACL-IJCNLP} and sentiment analysis \cite{DBLP:journals/corr/TaiSM15} to name a few.
The task to learn a distributed representation is called representation learning.

However, evaluating the quality of learned distributed word representation itself is not straightforward.
In language modeling, perplexity or cross-entropy is widely accepted as a de facto standard for intrinsic evaluation.
In contrast, distributed word representations include the additive (or compositional) property of the vectors, which cannot be assessed by perplexity.
Moreover, perplexity makes little use of infrequent words; thus, it is not appropriate for evaluating distributed presentations  that try to represent them.

Therefore, a word similarity task and/or a word analogy task are generally used to evaluate distributed word representations in the NLP literature.
The former judges whether distributed word representations improve modeling contexts, and the latter estimates how well the learned representations achieve the additive property.
However, such resources other than for English (e.g., Japanese) seldom exist.
In addition,  most of these datasets comprise high-frequency nouns so that they tend not to include other parts of speech.
Hence, previous data fail to evaluate word representations of other parts of speech, including content words such as verbs and adjectives.

To address the problem of the lack of a dataset for evaluating Japanese
distributed word representations, we propose to build a Japanese dataset for
the word similarity task.

The main contributions of our work are as follows:

\begin{itemize}
	\item To the best of our knowledge, it is the first work that constructs a Japanese word similarity dataset.
	\item The dataset contains various parts of speech and includes rare words in addition to common words.
\end{itemize}

\section{Related Work}

In general, distributed word representations are evaluated using a word similarity task.
For instance, WordSim353 \citelanguageresource{2002:PSC:503104.503110}, MC
\cite{strongContextualHypothesis}, RG
\cite{Rubenstein:1965:CCS:365628.365657}, and SCWS
\citelanguageresource{Huang:2012:IWR:2390524.2390645} have been used to evaluate word similarities in English.
Moreover, \newcite{baker-reichart-korhonen:2014:EMNLP2014}
built a verb similarity dataset (VSD) based on WordSim353 because there was no
dataset of verbs in the word-similarity task.
Recently, SimVerb-3500 was introduced to evaluate human understanding of verb
meaning \citelanguageresource{Gerz:2016:EMNLP}.
It provides human ratings for the similarity of 3,500 verb pairs so that
it enables robust evaluation of distributed representation for verbs.
However, most of these datasets include English words only. 
There has been no Japanese dataset for the word-similarity task.

Apart from English, WordSim353 and SimLex-999
\citelanguageresource{Hill:2015:CL} have been translated and
rescored in other languages: German, Italian and Russian
\citelanguageresource{Leviant:2015:arXiv}.
SimLex-999 has also been translated and rescored in Hebrew and Croatian
\citelanguageresource{Mrksic:2017:TACL}.
SimLex-999 explicitly targets at similarity rather than relatedness and
includes adjective, noun and verb pairs. However, this dataset contains
only frequent words.

In addition, the distributed representation of words is generally learned using only word-level information.
Consequently, the distributed representation for low-frequency words and unknown words cannot be learned well with conventional models.
However, low-frequency words and unknown words are often comprise high-frequency morphemes (e.g., unkingly $\rightarrow$ un + king + ly).
Some previous studies take advantage of the morphological information to provide a suitable representation for low-frequency words and unknown words \cite{Luong-etal:conll13:morpho,soricut-och:2015:NAACL-HLT}.
Morphological information is particularly important for Japanese since Japanese
is an agglutinative language.

\begin{table}[t]
  \begin{center}
    \begin{tabular}{ccccc} 
    \toprule
     Frequency & 1- & 101- & 1001- & 10001- \\
    \midrule
      Verb & 239 & 539 & 710 & 598  \\
      Adjective & 183 & 322 & 523 & 350 \\
      Noun & 15 & 63 & 172 & 258 \\
      Adverb & 23 & 75 & 80 & 81 \\
     \bottomrule
    \end{tabular}
    \caption{The number of parts of speech classified into each frequency.}
    \label{tab:freq}
  \end{center}
\end{table}

\section{Construction of a Japanese Word Similarity Dataset}

What makes a pair of words similar?
Most of the previous datasets do not concretely define the similarity of word pairs.
The difference in the similarity of word pairs originates from each annotator's mind, resulting in different scales of a word. 
Thus, we propose to use an example-based approach (Table \ref{tab:lancers}) to control the variance of the similarity ratings. 
We remove the context of word when we extracted the word.
So, we consider that an ambiguous word has high variance of the similarity, but we can get low variance of the similarity when the word is monosemous.
  
For this study, we constructed a Japanese word similarity dataset\footnote{\url{https://github.com/tmu-nlp/JapaneseWordSimilarityDataset}}.
We followed the procedure used to construct the Stanford Rare Word Similarity
Dataset (RW) \citelanguageresource{Luong-etal:conll13:morpho}.

We extracted Japanese word pairs from the Evaluation Dataset of Japanese
Lexical Simplification \citelanguageresource{kodaira}.
It targeted content words (nouns, verbs, adjectives, adverbs).
It included 10 contexts about target words annotated with their lexical substitutions and rankings.
Figure \ref{tab:example} shows an example of the dataset.
A word in square brackets in the text is represented as a target word of simplification.
A target word is not only recorded in the lemma form but also in the conjugated form.
We built a Japanese similarity dataset  from this dataset using the following procedure.

\paragraph{Word selection: }

First, paraphrase candidates were extracted from this dataset.
Because the construction process of the simplification dataset was divided into a paraphrase acquisition phase and a simplification ranking phase, we simply discarded the simplification rankings from the dataset to obtain paraphrase candidates.
Table \ref{tab:freq} shows the frequency of extracted words in the Japanese Wikipedia as of May 2015.
As shown in the table, low-frequency words are included in the dataset.

\paragraph{Pair construction: }

Because extracted words are annotated with their paraphrase candidates, we picked up each pair from the candidate as a word pair.
Consequently, we acquired 5,051 verb pairs, 4,033 adjective pairs, 1,528 noun pairs and 902 adverb pairs.
To balance the numbers of verb and adjective pairs with other parts of speech, we extracted samples at random for verbs and adjectives.
Finally, we obtained 1,464 verb pairs and 960 adjective pairs.

We observed that the similarity of the pairs extracted from the dataset of Kodaira et al. (2016) was low  without providing contexts; thus, we did not augment the dataset by inserting pseudo-negative instances from WordNet's synsets, as was done in the RW corpus.
Another reason why we did not employ the synset from the Japanese WordNet \cite{ISAHARA08.609} was because its quality was not as good as the English WordNet except for concrete nouns\footnote{It might be because it was translated from the English WordNet.
This is why we decided not to translate the existing English word similarity dataset to create a Japanese version.}.

\begin{table}
  \begin{center}
  \scalebox{0.75}{
    \begin{tabular}{cc|cc|c} 
    \toprule
      \multicolumn{2}{c|}{word 1} & \multicolumn{2}{c|}{word 2} & \multirow{2}{5mm}{sim.}  \\
      EN & JA & EN & JA  \\
     \midrule
      close & \begin{CJK}{UTF8}{min}瞑る\end{CJK} & close & \begin{CJK}{UTF8}{min}つぶる\end{CJK} & 10 \\
      erase & \begin{CJK}{UTF8}{min}拭き取る\end{CJK} & wipe & \begin{CJK}{UTF8}{min}拭う\end{CJK} & 8 \\
      mopey & \begin{CJK}{UTF8}{min}塞ぎ込んだ\end{CJK} & sick & \begin{CJK}{UTF8}{min}病んだ\end{CJK} & 5 \\
      investigate & \begin{CJK}{UTF8}{min}手探る\end{CJK} & go & \begin{CJK}{UTF8}{min}行く\end{CJK} & 2 \\
      fly & \begin{CJK}{UTF8}{min}とばせる\end{CJK} & control & \begin{CJK}{UTF8}{min}制御できる\end{CJK} & 0 \\
     \bottomrule
    \end{tabular}
    }
    \caption{Example of the degree of similarity when we requested annotation at Lancers.}
    \label{tab:lancers}
  \end{center}
\end{table}

\paragraph{Human judgment: }

We opted to use the crowd-sourcing service (Lancers\footnote{\url{http://www.lancers.jp}}) to hire native Japanese speakers.
We asked annotators to assign the degree of similarity for each pair using the
same 10-point scale\footnote{In a crowdsourcing request, we indicated that a
similarity of pairs with different notations, such as
``write\begin{CJK}{UTF8}{min}（書いた）\end{CJK}'' and
``write\begin{CJK}{UTF8}{min}（かいた）\end{CJK}'' is 10.}.
We used only those annotators who were able to complete at least 95\% of their previous assignments correctly.
We collected similarity rating for each word pair from ten annotators and defined the average of their annotations as the similarity of the pairs.

Although \citelanguageresource{kodaira} gave the annotators the context during annotation, we removed the context and gave only pairs to annotators.
We did so because the previous datasets such as VSD and RW did not present any context during annotation\footnote{Another reason why we did not do so is because the SCWS has a very high variance even though it is annotated with contexts (Table 5).}.
To improve the quality of the annotation, we presented an example of the degree of similarity of the pairs during annotation (Table \ref{tab:lancers}).
Consequently, we collected 4,851 pairs overall.
Table \ref{tab:lanexample} shows an example of a pair from our dataset.
Inter-annotator agreements (IAA) of each POS are shown in Table \ref{tab:IAA}.   
The inter-annotator agreement is the average Spearman's $\rho$ between a single annotator and the average of all others.

\begin{table}
  \begin{center}
    \begin{tabular}{ccccc} 
    \toprule
        POS & verb & adj & adv & noun \\
     \midrule
	IAA & 0.69 & 0.67 & 0.61 & 0.56 \\
     \bottomrule
    \end{tabular}
    \caption{Inter-annotator agreements of each POS.}
    \label{tab:IAA}
  \end{center}
\end{table}

\begin{table*}[htbp]
  \begin{center}
    \begin{tabular}{p{8mm}p{3mm}cccccc} 
    \toprule
      \multirow{2}{13mm}{word 1} & EN & follow & exclude & challenge & storm & elucidate &  wander \\
                                           & JA  & \begin{CJK}{UTF8}{min}受け継ぐ\end{CJK} & \begin{CJK}{UTF8}{min}除外する\end{CJK} & \begin{CJK}{UTF8}{min}チャレンジする\end{CJK} & \begin{CJK}{UTF8}{min}しける\end{CJK} & \begin{CJK}{UTF8}{min}明白になる\end{CJK} & \begin{CJK}{UTF8}{min}迷う\end{CJK} \\
      \multirow{2}{13mm}{word 2} & EN & inherit & remove & wish & rough & reflect &  stop \\
                                           & JA  & \begin{CJK}{UTF8}{min}継承する\end{CJK} & \begin{CJK}{UTF8}{min}除去する\end{CJK} & \begin{CJK}{UTF8}{min}望む\end{CJK} & \begin{CJK}{UTF8}{min}あれる\end{CJK} & \begin{CJK}{UTF8}{min}反映される\end{CJK} & \begin{CJK}{UTF8}{min}止める\end{CJK} \\ 
      \multicolumn{2}{c}{similarity}  & 9.3 & 7.3 & 6.0 & 5.7 & 2.7 & 1.7 \\
     \bottomrule
    \end{tabular}
    \caption{Examples of verb pairs in our dataset.
     The similarity rating is the average of the ratings from ten annotators. }
    \label{tab:lanexample}
  \end{center}
\end{table*}

\section{Discussion}

\subsection{Comparison to Other Datasets}

Table \ref{tab:result} shows how several resources vary.
WordSim353 comprises high-frequency words and so the variance tends to be low.
In contrast, RW includes low-frequency words, unknown words, and complex words composed of several morphemes; thus, the variance is large.
VSD has many polysemous words, which increase the variance.
Despite the fact that our dataset, similar to the VSD and RW datasets, contains low-frequency and ambiguous words, its variance is 3.00.
The variance level is low compared with the other corpora.
We considered that the examples of the similarity in the task request reduced the variance level.

\begin{table}[t]
  \begin{center}
    \begin{tabular}{cc} 
    \toprule
      Dataset & Variance  \\
     \midrule
      WordSim353 & 3.16 \\
      VSD & 4.76 \\
      RW  & 5.70 \\
      SCWS  & 8.60 \\
      JWSD (our dataset) & 3.00 \\
     \bottomrule
    \end{tabular}
    \caption{Variance of each dataset.}
    \label{tab:result}
  \end{center}
\end{table}

We did not expect SCWS to have the largest variance in the datasets shown in Table \ref{tab:result} because it gave the context to annotators during annotation.
At the beginning, we thought the context would serve to remove the ambiguity and clarify the meaning of word; however after looking into the dataset, we determined that the construction procedure used several extraordinary annotators.
It is crucial to filter insincere annotators and provide straightforward instructions to improve the quality of the similarity annotation like we did.

To gain better similarity, each dataset should utilize the reliability score to exclude extraordinary annotators.
For example, for SCWS, an annotator rating the similarity of pair of  ``CD'' and ``aglow'' assigned a rating of 10.
We  assumed it was a typo or misunderstanding regarding the words.
To address this problem, such an annotation should be removed before calculating the true similarity.
All the datasets except for RW simply calculated the average of the similarity, but datasets created using crowdsourcing should consider the reliability of the annotator.

\subsection{Analysis}

We present examples of a pair with high variance of similarity as shown below:

\paragraph{Aspect of relatedness.} (e.g., a pairing of
``fast\begin{CJK}{UTF8}{min}（速い）\end{CJK}'' and
``early\begin{CJK}{UTF8}{min}（早い）\end{CJK}''.) 

Although they are similar in meaning with respect to the time, they have
nothing in common with respect to speed; Annotator A assigned a rating of 10,
but Annotator B assigned a rating of 1.

Another example, the pairing of ``be
eager\mbox{\begin{CJK}{UTF8}{min}（懇願する）\end{CJK}''} and ``request\begin{CJK}{UTF8}{min}（頼む）\end{CJK}''.
Even though the act indicated by the two verbs is the same, there are some cases where they express different degrees of feeling.
Compared with ``request'', ``eager'' indicates a stronger feeling.
There were two annotators who emphasized the similarity of the act itself
rather than the different degrees of feeling, and vice versa.
In this case, Annotator A assigned a rating of 9, but Annotator B assigned a rating of 2.

Although it was necessary to distinguish similarity and semantic relatedness
\cite{Mrksic:2016:NAACL} and we asked annotators to rate the pairs based on
semantic similarity, it was not straightforward to put paraphrase candidates
onto a single scale considering all the attributes of the words.
This limitation might be relaxed if we would ask annotators to refer to a
thesaurus or an ontology such as Japanese Lexicon
\citelanguageresource{GoiTaikei:1997}.

\paragraph{Comparing spell\footnote{We indicated these pair's similarity is 10. However, some annotators ignored this instruction. It would be necessary to clean the spellings of paraphrase candidates before requesting similarity annotation.}.} (e.g., a pairing of ``slogan\begin{CJK}{UTF8}{min}（スローガン）\end{CJK}'' and ``slogan\begin{CJK}{UTF8}{min}（標語）\end{CJK}''.) 

In Japanese, we can write a word using hiragana, katakana, or kanji characters; however because hiragana and katakana represent only the pronunciation of a word, annotators might think of different words. 
In this case, Annotator A assigned a rating of 8, but Annotator B assigned a rating of 0.
Similarly, we confirmed the same thing in other parts of speech.
Especially, nouns can have several word pairs with different spellings, which
results in their IAA became too low compared to other parts of speech.

\paragraph{Frequency or time expressions.} (e.g.,  a pairing of ``often\begin{CJK}{UTF8}{min}（しばしば）\end{CJK}'' and ``frequently\begin{CJK}{UTF8}{min}（しきりに）\end{CJK}''.)

We confirmed that the variance becomes larger among adverbs expressing frequency.
This is due to the difference in the frequency of words that annotators imagines.
In this case, Annotator A assigned a rating of 9, but Annotator B assigned a rating of 0.
Similarly, we confirmed the same thing among adverbs expressing time.

\section{Conclusion}

In this study, we constructed the first Japanese word similarity dataset.
It contains various parts of speech and includes rare words in addition to common words.
Crowdsourced annotators assigned similarity to word pairs during the word similarity task.
We gave examples of similarity in the task request sent to annotators, so that we reduced the variance of each word pair.
However, we did not restrict the attributes of words, such as the level of feeling, during annotation.
Error analysis revealed that the notion of similarity should be carefully defined when constructing a similarity dataset.

As a future work, we plan to construct a word analogy dataset in Japanese by translating an English dataset to Japanese.
We hope that a Japanese database will facilitate research in Japanese distributed representations.

\section{Bibliographical References}
\bibliographystyle{lrec}
\bibliography{lrec2018}

\section{Language Resource References}
\bibliographystylelanguageresource{lrec}
\bibliographylanguageresource{lrec2018}

\end{document}